\documentclass[conference]{IEEEtran}
\IEEEoverridecommandlockouts
\usepackage{cite}
\usepackage{amsmath,amssymb,amsfonts}
\usepackage{algorithm}
\usepackage{algorithmicx}
\usepackage{algpseudocode}
\algrenewcommand\algorithmicrequire{\textbf{Input:}}
\algrenewcommand\algorithmicensure{\textbf{Output:}}

\usepackage[english]{babel}
\usepackage{graphicx}
\usepackage{textcomp}
\usepackage{xcolor}
\usepackage{soul}
\usepackage{multirow}
\usepackage{array}
\usepackage{hhline}
\usepackage{titlesec}
\usepackage{authblk}
\usepackage{textcomp}
\usepackage{stfloats}
\usepackage{url}
\usepackage{verbatim}
\usepackage{tikz}
\mathchardef\mhyphen="2D
\usepackage{color}
\usepackage{microtype}
\usepackage{wrapfig}
\usepackage{booktabs} 
\usepackage{dsfont}
\usepackage{pgfplots}
\usetikzlibrary{patterns}
\usepgfplotslibrary{fillbetween}
\usepackage{pgfkeys}
\usepackage{lipsum}
\usepackage{mwe}
\usepackage{tikz}
\usetikzlibrary{backgrounds}
\usepackage[style=base]{caption}
\usetikzlibrary{intersections}
\usetikzlibrary{shapes.geometric}
\usetikzlibrary{spy}
\usetikzlibrary{shapes.geometric}
\usepackage{pgfplotstable}
\pgfmathdeclarefunction{gauss}{2}{%
  \pgfmathparse{1/(#2*sqrt(2*pi))*exp(-((x-#1)^2)/(2*#2^2))}%
}
\pgfplotsset{compat=1.11,
    /pgfplots/ybar legend/.style={
    /pgfplots/legend image code/.code={%
       \draw[##1,/tikz/.cd,yshift=-0.25em]
        (0cm,0cm) rectangle (3pt,0.8em);},
   },
}
\usepackage{colortbl}
\usepgfplotslibrary{colormaps}
    \def\addlegendimage{\csname pgfplots@addlegendimage\endcsname}

\pgfplotsset{ 
cycle list={%
{draw=black,mark=star,solid},
{draw=black, mark=square,solid}}}
\usepackage{breqn}

\newcommand{\brokenline}[2][t]{\parbox[#1]{\dimexpr\linewidth-\ALG@thistlm}{\strut\raggedright #2\strut}}

\renewcommand{\fnum@figure}{Fig. \thefigure}

\usepackage[a4paper, total={187mm,250mm}]{geometry}
\def\BibTeX{{\rm B\kern-.05em{\sc i\kern-.025em b}\kern-.08em
    T\kern-.1667em\lower.7ex\hbox{E}\kern-.125emX}}

\newcommand\blfootnote[1]{%
  \begingroup
  \renewcommand\thefootnote{}\footnote{#1}%
  \addtocounter{footnote}{-1}%
  \endgroup
}

\usepackage{caption}
\usepackage{fancyhdr}
\fancypagestyle{firstpage}
{
    \fancyhead[L]{\footnotesize © 2023 IEEE.  Personal use of this material is permitted.  Permission from IEEE must be obtained for all other uses, in any current or future media, including reprinting/republishing this material for advertising or promotional purposes, creating new collective works, for resale or redistribution to servers or lists, or reuse of any copyrighted component of this work in other works. This paper is accepted at the IEEE International Conference on Artificial Intelligence Circuits and Systems (AICAS) 2023.}
    \fancyhead[R]{}
}

\begin{document}

\title{\LARGE \textbf{Enhancing Fault Resilience of QNNs by Selective Neuron Splitting} \vspace{-3mm}}

\author[1]{Mohammad Hasan Ahmadilivani}
\author[1]{Mahdi Taheri}
\author[1]{Jaan Raik}
\author[1,2]{Masoud Daneshtalab}
\author[1]{Maksim Jenihhin \vspace{-2mm}}
\affil[1]{Tallinn University of Technology, Tallinn, Estonia}
\affil[2]{Mälardalen University, Västerås, Sweden}
\affil[1]{\{mohammad.ahmadilivani, mahdi.taheri, jaan.raik, maksim.jenihhin\}@taltech.ee}
\affil[2]{masoud.daneshtalab@mdu.se \vspace{-6mm}}

\maketitle
\thispagestyle{firstpage}

\begin{abstract}

The superior performance of Deep Neural Networks (DNNs) has led to their application in various aspects of human life. Safety-critical applications are no exception and impose rigorous reliability requirements on DNNs. Quantized Neural Networks (QNNs) have emerged to tackle the complexity of DNN accelerators, however, they are more prone to reliability issues.

In this paper, a recent analytical resilience assessment method is adapted for QNNs to identify critical neurons based on a Neuron Vulnerability Factor (NVF). Thereafter, a novel method for splitting the critical neurons is proposed that enables the design of a Lightweight Correction Unit (LCU) in the accelerator without redesigning its computational part.

The method is validated by experiments on different QNNs and datasets. The results demonstrate that the proposed method for correcting the faults has a twice smaller overhead than a selective Triple Modular Redundancy (TMR) while achieving a similar level of fault resiliency. 

\end{abstract}

\blfootnote{The work is supported in part by the EU through European Social Fund in the frames of the ``ICT programme'' (``ITA-IoIT'' topic), by the Estonian Research Council grant PUT PRG1467 ``CRASHLESS'', Estonian Centre for Research Excellence EXCITE and by Estonian-French PARROT project ``EnTrustED''.}


\vspace*{-5mm}

\section{Introduction}  \label{sec:intro}

Artificial Intelligence (AI) has shifted the paradigm of computer science in the latest decade with Deep Neural Networks (DNNs), one of AI's illustrious instruments, demonstrating remarkable precision levels \cite{silver2017mastering}. This has led to their adoption in several safety-critical applications like autonomous driving \cite{mozaffari2020deep}. As DNN accelerators become more prevalent in safety-critical applications, hardware reliability of digital circuits has become increasingly more noticeable. The reliability of DNNs is determined by the ability of their accelerators to function correctly \cite{ibrahim2020soft} in the presence of environment-related faults (soft errors, electromagnetic effects, temperature variations) or faults in the underlying hardware (manufacturing defects, process variations, aging effects) \cite{shafique2020robust}. 

Various emerging techniques are explored to improve the computational efficiency of DNNs' complex architectures, such as reducing the bit precision of parameters, which has led to the emergence of Quantized Neural Networks (QNNs). However, the effectiveness of such techniques raises concerns about the reliability of QNNs, particularly in safety-critical applications. Soft errors, a type of fault caused by charged particles colliding with transistors, can cause a logic value to flip, dramatically influencing the functionality of QNNs \cite{zahid2020fat,khoshavi2020fiji}.

Throughout the literature, protecting DNNs against soft errors is primarily achieved through architecture-level methods such as hardened PEs or Triple Modular Redundancy (TMR) \cite{mittal2020survey}. However, to alleviate overheads, there is a need, first, to identify the critical neurons within a neural network before applying the mentioned mitigation techniques to harden them against the faults.

Reliability assessment serves as the initial step towards exploiting an effective protection mechanism. Fault Injection (FI) is a conventional method for reliability assessment that is vastly adopted for DNNs. However, identifying the critical points in a QNN requires an exhaustive FI that is prohibitively complex due to their large number of parameters. To address this issue, analytical resilience assessment approaches are proposed to evaluate the reliability of DNNs by analyzing them at the algorithm level \cite{mahmoud2020hardnn}. 

In previous works, the criticality of neurons has been identified based on their contribution scores to outputs \cite{schorn2018accurate,schorn2019efficient,ruospo2021reliability,abdullah2020salvagednn}. Hence, there is no clear resilience evaluation metric for selecting the critical neurons in the literature, and recent works extract the criticality based on the ranked scores. To tackle the drawbacks of the state-of-the-art in DNNs' resilience analysis methods, a prior study has proposed a method called DeepVigor \cite{deepvigor}, which provides vulnerability factors for all bits, neurons, and layers of DNNs accurately. However, it does not consider QNNs. In this work, we adapt and optimize DeepVigor for identifying critical neurons in QNNs. The resilience analysis enables us to design a method for correcting soft errors in the datapath of DNN accelerators.

In this paper, we identify critical neurons in QNNs based on a Neuron Vulnerability Factor (NVF) obtained by fault propagation analysis through the QNNs. The NVF represents the probability of misclassification due to a fault in a neuron which determines the level of criticality for neurons. To the best of our knowledge, for the first time, a protection technique based on splitting neurons' operations is proposed that modifies the network in a way that a Lightweight Correction Unit (LCU) corrects the faults in critical neurons. The proposed method does not require redesigning the computational part of the accelerator. The accelerator executes the modified network, and only its controller needs to be aware of the critical neurons to be operated on the LCU. Our method imposes half the overhead of TMR since it corrects faults with only one additional neuron instead of two.

The contributions of this work are as follows:

\begin{itemize}
    \item Developing an analytical fault resilience assessment method for QNNs to identify the most critical neurons based on the conducted Neuron Vulnerability Factor (NVF);
    \item Proposing a novel high-level modification method for QNNs to improve fault resiliency by splitting the operations of critical neurons, without requiring a redesign of the computational part of the  accelerator;
    \item Designing an effective Lightweight Correction Unit (LCU) for selected critical neurons in accelerators, with low overhead (twice less than that of TMR) and high fault resiliency (similar to that of TMR).
\end{itemize}

The paper is organized as follows. The proposed method for enhancing fault resilience of QNNs is presented in Section \ref{sec:method}, experiments are performed and discussed in Section \ref{sec:experiments}, and the paper is concluded in Section \ref{sec:conclusion}.

\section{Method For Resilience Enhancement of QNNs}   \label{sec:method}

\subsection{Accelerator Model} \label{method:accelerator}

\begin{figure}[h]
    \includegraphics[width=0.35\textwidth]{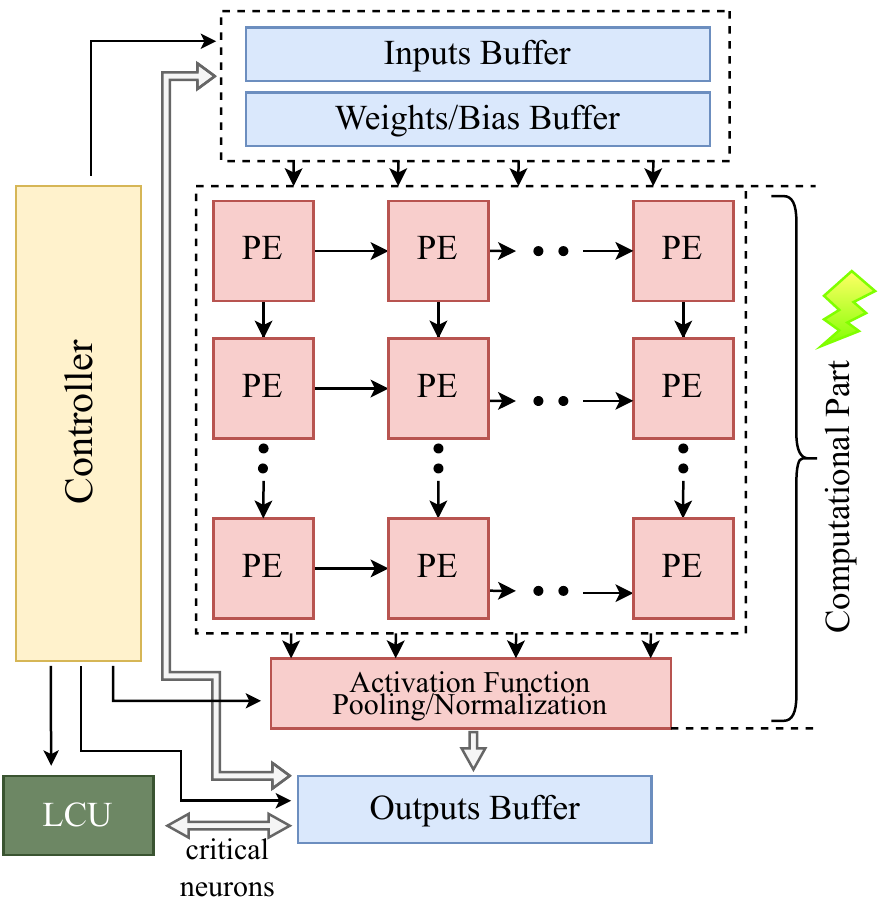}
    \centering
    \caption {An abstract view of the accelerator and where the faults may happen.}
    \label{fig:acc-model}
\end{figure}

Fig. \ref{fig:acc-model} illustrates the accelerator model considered in this work which is inspired by \cite{ozen2020just}. It consists of a computational part (an array of Processing Elements (PEs), activation functions, pooling, and normalization), buffers for parameters (weight and bias), inputs, and outputs, and the controller. It is assumed that faults may happen in the computational part of the accelerator, thus, the \textit{Outputs Buffer} may contain faulty values of output activations. The controller is responsible for feeding the inputs, transferring the outputs, and controlling the function of the accelerator. 

To apply the resilience enhancement method to accelerator, a Lightweight Correction Unit (LCU) is added to the design in which the controller only needs to be aware of the critical neurons. Once the outputs of a layer are calculated, the controller transfers the critical neurons to LCU, replaces its corrected outputs back to the \textit{Outputs Buffer}, and continues the operations of the accelerator. The design of the LCU is proposed in Subsection \ref{method:techniques}.

\subsection{Identifying Critical Neurons by Resilience Analysis} \label{method:analysis}

Algorithm \ref{alg:resilience-analysis} presents the resilience analysis of QNNs to obtain Neuron Vulnerability Factors (NVF) for all neurons throughout the QNN in convolution and fully-connected layers. It is assumed that the neural network is quantized into an 8-bit signed integer data type, and the output activation of the neuron is analyzed. The algorithm, first, checks whether or not to analyze an input for the neuron (lines 3-5) by the gradients of a loss function ($\mathcal{L}$) that represents the impact of the neuron's erroneous output on the golden top class of the network.

Then, it finds minimum positive and maximum negative values for the neuron ($\delta$), that cause a misclassification in the QNN from its golden output (lines 6, 7). Thereafter, it maps the obtained $\delta$ to a corresponding possible bitflip location in the data type (lines 8, 9) and counts it as a vulnerable location (lines 10, 11). In the end, regarding the counted of vulnerable times for each bit, it calculates the probability of misclassification of the network by each bitflip in the output of the neuron as the NVF over the whole inputs (line 15).

A key observation in the analysis is that the \textit{0 to 1} bitflip is much more critical than \textit{1 to 0} bitflip. Because the former enlarges the values in the activation and propagates to the output, while the latter is masked. This observation leads us to the protection mechanism proposed in the next Subsection. It is worth mentioning that the resilience analysis method is not limited to a single-bit flip fault model, and it implicitly considers multi-bit faults.

By obtaining the NVF of all neurons through the QNN, the critical neurons can be found based on the values for NVF. Different thresholds can be set to select the critical neurons and protect them, considering how many of them are affected by the protection techniques leading to execution overheads.

\begin{algorithm}[hbpt]
    \caption{Resilience Analysis of QNNs}
    \label{alg:resilience-analysis}
    \begin{algorithmic}[1]
\Require{Trained QNN with a set of neurons \textit{Q} and \textit{N} outputs, set of input images \textit{X}};
\Ensure{NVF of all neurons};
\Statex Assume: $\delta \in $ [-128,127]; $\mathcal{E}_{c_t}$ is the output score for the golden top class; $C_{g}$ is golden classification; $C_{\delta}$ is classification result after injecting $\delta$; \textit{vul\_map\_arr\_pos} and \textit{vul\_map\_arr\_neg} include counters for each bit corresponds to each vulnerability range for positive and negative numbers;
    \For{$neuron \in $ Q}:
        \For {$input \in X$}:
            \State \brokenline{%
            $\mathcal{L} = sigmoid(\sum_{j=0}^{N}(\mathcal{E}_{c_t} - \mathcal{E}_{c_j}))$}
            \State \brokenline{%
            \textit{grad} = $\nabla \mathcal{L} / out_{neuron}$}
            \If{\textit{grad} != 0}
            \State \brokenline{%
             $r_{upper} = min(\delta), \delta>0 , s.t. \; C_{g} \neq C_{f}$}
             \State \brokenline{%
             $r_{lower} = max(\delta), \delta<0 , s.t. \; Cl_{g} \neq Cl_{f}$}
            \State \brokenline{%
            $bit_{upper}$ = int($\sqrt{r_{upper}}$) + 1;}
            \State \brokenline{%
            $bit_{lower}$ = int($\sqrt{|r_{lower}|}$);}
            \State \brokenline{%
            $vul\_map\_arr\_pos[bit_{upper}]$++;}
            \State \brokenline{%
            $vul\_map\_arr\_neg[bit_{lower}]$++;}
            \EndIf;
        \EndFor;
        \State \brokenline{%
        $vul\_map\_arr =  (vul\_map\_arr\_pos + vul\_map\_arr\_neg) \; / \; 2$}
        \State \brokenline{%
        $NVF_{neuron} = \frac{\sum_{i=1}^{8} (\frac{1}{8} \times \sum_{j=1}^i(vul\_map\_arr[j]))}{size(X)}$}
    \EndFor;
    \end{algorithmic}
\end{algorithm}

\subsection{Resilience Enhancement by Splitting Critical Neurons and LCU} \label{method:techniques}

\begin{figure}[h]
    \includegraphics[width=0.4\textwidth]{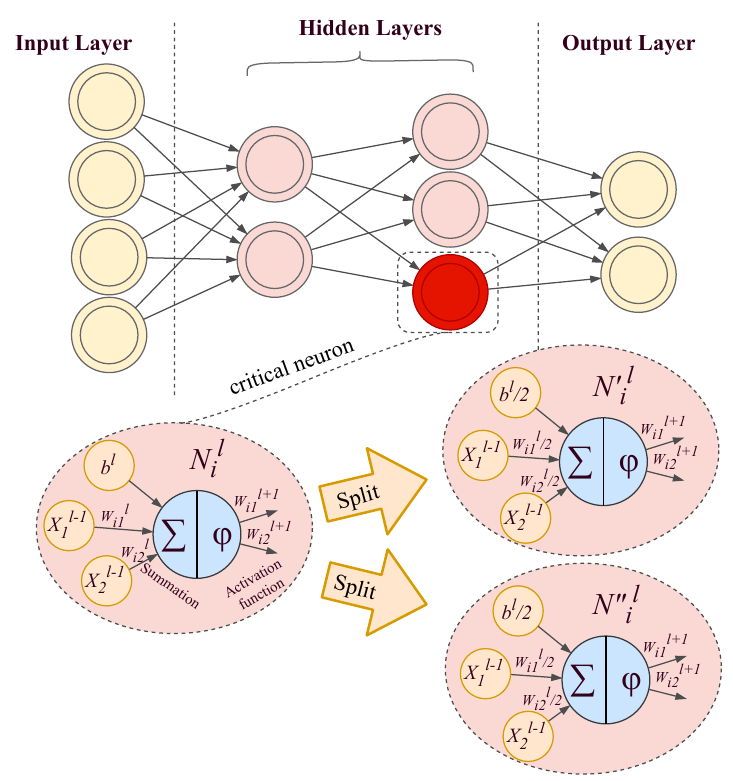}
    \centering
    \caption {Operation splitting for a neuron in a QNN involves halving the input parameters while keeping the output parameters non-modified. A critical neuron is replaced with its corresponding split neurons in the QNN.}
    \label{fig:neuron-split}
\end{figure}

The proposed fault resilience enhancement targets the critical neurons identified based on a threshold on NVF. The idea is to split the selected neurons' operation into two neurons in the QNN at a high level and correct the critical outputs in the accelerator. Fig. \ref{fig:neuron-split} depicts how a critical neuron is split into two halves. As it is shown, the input parameters (weights and bias) of the neuron are halved, keeping the output parameters non-modified, and the new neurons are replaced with the critical neuron in the QNN. In this way, the neuron can be split into two neurons without changing the intermediate values of the further layers and the neural network's outputs. Noteworthy, the method is applied to all identified critical neurons in convolution and fully-connected layers.

Splitting the critical neurons provides an opportunity for fault correction using the split neurons without redesigning the computational part of the accelerator. The network is modified in a way that the selected critical neurons from the analysis are split. The modified network can then be mapped to the accelerator using the existing controller and mapping algorithm of the accelerator. However, the controller needs to be aware of the critical neurons so that it can transfer them to LCU to perform the correction and write them back to the \textit{Output Buffers} (Fig. \ref{fig:acc-model}).

LCU is designed to leverage the neuron-splitting method for correction. The inputs of LCU are two split neurons representing one critical neuron, and the output is one corrected 8-bit data that will be written back to the corresponding neurons. 

The data type (signed integer 8-bit) contains one sign bit and 7 bits for the integer. As the neuron's operation is split, the range of output values for each replaced neuron would be divided by 2. Therefore, the Most Significant Bit (MSB) in the integer part of the output should always be 0. Regarding the observation in the analysis about bitflips (Subsection \ref{method:analysis}), any faulty bit can be set to zero to be less critical. 

\begin{figure}[h]
    \includegraphics[width=0.4\textwidth]{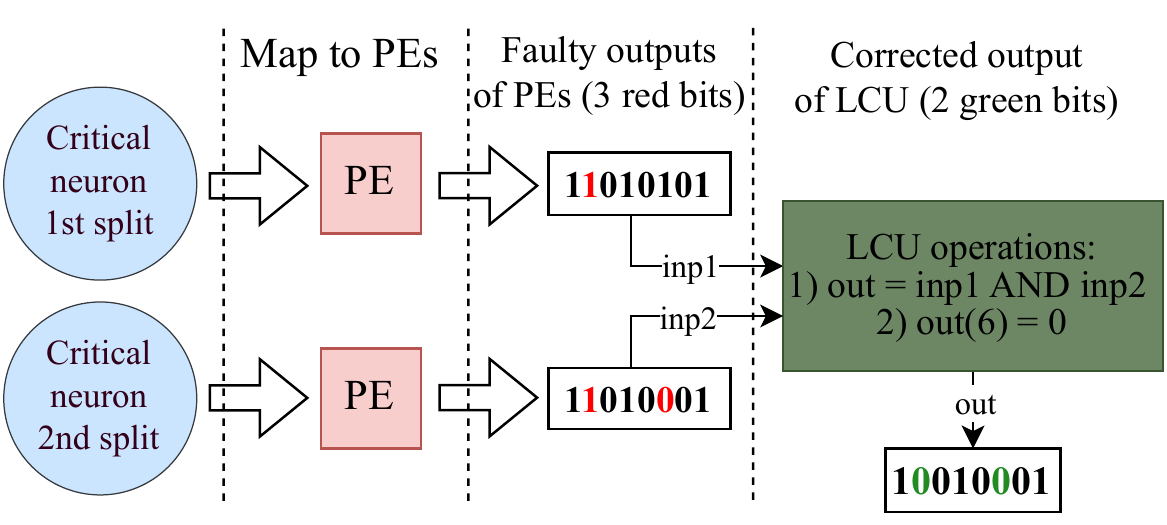}
    \centering
    \caption{An example of how LCU corrects faulty critical neurons.}
    \label{fig:lcu-op}
\end{figure}

Therefore, to output the corrected value, LCU performs two operations: 1) a bit-wise AND over the two inputs, 2) resets the MSB of the integer part to 0. In this way, many single and also multiple faults that occur to the bits will be masked by these two operations. Since the correction operations are merely an \textit{AND} and a \textit{bit reset}, the correction unit is \textit{lightweight}. The operation of the LCU correction is depicted in Fig.~\ref{fig:lcu-op} performing on the faulty outputs of PEs running two splits of a critical neuron. The corrected output is written back to \textit{Outputs Buffer} as the outputs of the corresponding PEs. 

\section{Experiments} \label{sec:experiments}

\subsection{Experimental Setup}

The experimented QNNs in this work are fully quantized (all parameters and activation) to 8-bit signed integer using TFLite \cite{david2021tensorflow}. The experiments in this work have been performed on a 7-layer MLP and LeNet-5 trained on MNIST as well as an AlexNet trained on CIFAR-10. The baseline accuracy of each network on the test data is 70.1\%, 89.1\%, and 62.9\%, respectively.

The resilience analysis and enhancement (Sections \ref{method:analysis} and \ref{method:techniques}) are implemented in PyTorch considering the accelerator model. The resilience analysis is conducted over the training set. The critical neurons regarding different thresholds for NVF are obtained to explore the number of neurons to be protected, which imposes an overhead as well. 

To show the efficacy of the resilience enhancement method, a statistical FI is performed. In the FI process, one single bitflip in the output of a random neuron in the network is injected, and whole inference over the test set is performed, and the overall accuracy is obtained. To meet the 95\% confidence level with a 1\% error margin in the statistical FI based on \cite{leveugle2009statistical}, we repeated the FI process for each MLP-7, LeNet-5, and AlexNet for 6,750, 7,650, and 9,500 random faults, respectively.

As a baseline comparison of the proposed design for LCU, we also apply a TMR to the critical neurons for the detection and correction of faults. We adopt two metrics for comparing the results of methods and expressing the resiliency: 1) accuracy loss of QNNs over the fault injection, 2) the portion of critical faults in a fault injection campaign. Critical faults are the ones that misclassify the network from its golden classification. 

\subsection{Experimental Results}

\subsubsection{An Exploration on NVF of QNNs}

As mentioned, NVF explores the probability of a faulty neuron's output that misclassifies the QNN from its golden output. Table \ref{tab:res-nvf-exp} presents the number of critical neurons in different NVFs ranging from 0\% (all neurons are critical) to 50\% (no neuron is critical). According to the table, different thresholds of NVF count a different portion of neurons as critical among QNNs. However, it is observed that all neurons among QNNs have NVF of less than 50\%. It is noteworthy that a higher threshold for NVF means a less number of critical neurons to be protected. This table represents the overhead of any protection mechanism over the critical neurons.

\begin{table}[h]
\small
\centering
\caption{Exploration of number and portion of critical neurons over different thresholds for NVF.}
\resizebox{\columnwidth}{!}{%
\begin{tabular}{|c||cc||cc||cc|}
\hline
QNN                      & \multicolumn{2}{c||}{MLP-7}             & \multicolumn{2}{c||}{LeNet-5}           & \multicolumn{2}{c|}{AlexNet}            \\ \hline
NVF threshold            & \multicolumn{1}{c|}{\#neurons} & portion & \multicolumn{1}{c|}{\#neurons} & portion & \multicolumn{1}{c|}{\#neurons} & portion \\ \hline
NVF \textgreater{}= 0\%  & \multicolumn{1}{c|}{2816}     & 100\%   & \multicolumn{1}{c|}{4684}     & 100\%   & \multicolumn{1}{c|}{103168}   & 100\%   \\ \hline
NVF \textgreater{}= 5\%  & \multicolumn{1}{c|}{2513}     & 89.24\% & \multicolumn{1}{c|}{4380}     & 93.5\%  & \multicolumn{1}{c|}{46322}    & 44.9\%  \\ \hline
NVF \textgreater{}= 10\% & \multicolumn{1}{c|}{1382}     & 49.07\% & \multicolumn{1}{c|}{1659}     & 35.41\% & \multicolumn{1}{c|}{15818}    & 15.33\% \\ \hline
NVF \textgreater{}= 15\% & \multicolumn{1}{c|}{903}      & 32.06\% & \multicolumn{1}{c|}{222}      & 4.74\%  & \multicolumn{1}{c|}{5171}     & 5.01\%  \\ \hline
NVF \textgreater{}= \textbf{20\%} & \multicolumn{1}{c|}{\textbf{503}}      & \textbf{17.86\%} & \multicolumn{1}{c|}{\textbf{187}}      & \textbf{3.99\%}  & \multicolumn{1}{c|}{\textbf{622}}      & \textbf{0.6\%}   \\ \hline
NVF \textgreater{}= 25\% & \multicolumn{1}{c|}{272}      & 9.6\%   & \multicolumn{1}{c|}{70}       & 1.49\%  & \multicolumn{1}{c|}{398}      & 0.38\%  \\ \hline
NVF \textgreater{}= 30\% & \multicolumn{1}{c|}{184}      & 6.5\%   & \multicolumn{1}{c|}{3}        & 0.06\%  & \multicolumn{1}{c|}{232}      & 0.2\%   \\ \hline
NVF \textgreater{}= 35\% & \multicolumn{1}{c|}{85}       & 3.01\%  & \multicolumn{1}{c|}{0}        & 0\%     & \multicolumn{1}{c|}{147}      & 0.14\%  \\ \hline
NVF \textgreater{}= 40\% & \multicolumn{1}{c|}{26}       & 0.92\%  & \multicolumn{1}{c|}{0}        & 0\%     & \multicolumn{1}{c|}{56}       & 0.05\%  \\ \hline
NVF \textgreater{}= 45\% & \multicolumn{1}{c|}{7}        & 0.2\%   & \multicolumn{1}{c|}{0}        & 0\%     & \multicolumn{1}{c|}{6}        & 0.005\% \\ \hline
NVF \textgreater{}= 50\% & \multicolumn{1}{c|}{0}        & 0\%     & \multicolumn{1}{c|}{0}        & 0\%     & \multicolumn{1}{c|}{0}        & 0\%     \\ \hline
\end{tabular}%
}
\label{tab:res-nvf-exp}
\end{table}

\subsubsection{Resilience Enhancement of QNNs}

\begin{table*}[ht!]
\captionsetup{justification=centering}
\captionsetup{justification=centering}
\centering
\resizebox{2\columnwidth}{!}{
\begin{tabular}{ccc}

\begin{tikzpicture}
\pgfplotsset{
  x ticks with fixed point/.style={
      xticklabel={
        \pgfkeys{/pgf/fpu=true}
        \pgfmathparse{exp(\tick)}%
        \pgfmathprintnumber[fixed relative, precision=3]{\pgfmathresult}
        \pgfkeys{/pgf/fpu=false}
      }
  },
  y ticks with fixed point/.style={
      yticklabel={
        \pgfkeys{/pgf/fpu=true}
        \pgfmathparse{exp(\tick)}%
        \pgfmathprintnumber[fixed relative, precision=3]{\pgfmathresult}
        \pgfkeys{/pgf/fpu=false}
      }
  }
}
  \begin{axis}[title=MLP-7 (MNIST),
        width=0.35\textwidth,
        height=0.5\columnwidth,
        scaled x ticks = false,
        scaled y ticks = false,
        xtick={0,5,10,15,20,25,30,35,40,45,50}, 
        ytick={0,5,10,15,20},
        xticklabels = {\strut  $0$,\strut $5$,\strut $10$,\strut $15$,\strut $20$,\strut $25$, \strut$30$,\strut$35$,\strut$40$,\strut$45$, \strut$50$},
        yticklabels = {\strut $0$, \strut $5$,\strut $10$,\strut $15$,\strut $20$},
        ymin=0, ymax=20,
        xmin =0, xmax=50,
        grid=major, 
        grid style={dashed,gray}, 
        ylabel near ticks,
        xlabel near ticks,
        xlabel= NVF (\%), 
        ylabel= Accuracy Loss (\%),
        legend columns = 3,
        legend style={at={(1,1)},anchor=north},
        legend style={draw=black, at={(0.5,1)}, text opacity = 1,row sep=0pt, font=\fontsize{6}{4}\selectfont},
         x tick label style={rotate=0,anchor=north},
        ]
        \addplot [smooth, black, thick ,mark=square , mark size=2pt] table [x=NVF, y=Unprotected, col sep=comma] {images/quantization-mlp7-acc-loss.csv};
        \addplot [smooth, blue, thick ,mark=o , mark size=2pt] table [x=NVF, y=Proposed, col sep=comma] {images/quantization-mlp7-acc-loss.csv};
         \addplot [smooth, red, thick,mark=triangle*,mark size=2pt] table [x=NVF, y=TMR, col sep=comma] {images/quantization-mlp7-acc-loss.csv};
    
        \legend{Unprotected, Proposed, TMR}
      \end{axis}
    \end{tikzpicture}

&

\begin{tikzpicture}
\pgfplotsset{
  x ticks with fixed point/.style={
      xticklabel={
        \pgfkeys{/pgf/fpu=true}
        \pgfmathparse{exp(\tick)}%
        \pgfmathprintnumber[fixed relative, precision=3]{\pgfmathresult}
        \pgfkeys{/pgf/fpu=false}
      }
  },
  y ticks with fixed point/.style={
      yticklabel={
        \pgfkeys{/pgf/fpu=true}
        \pgfmathparse{exp(\tick)}%
        \pgfmathprintnumber[fixed relative, precision=3]{\pgfmathresult}
        \pgfkeys{/pgf/fpu=false}
      }
  }
}
  \begin{axis}[title=LeNet-5 (MNIST),
        width=0.35\textwidth,
        height=0.5\columnwidth,
        scaled x ticks = false,
        scaled y ticks = false,
        xtick={0,5,10,15,20,25,30,35,40,45,50}, 
        ytick={0,5,10,15, 20},
        xticklabels = {\strut  $0$,\strut $5$,\strut $10$,\strut $15$,\strut $20$,\strut $25$, \strut$30$,\strut$35$,\strut$40$,\strut$45$, \strut$50$},
        yticklabels = {\strut $0$, \strut $5$,\strut $10$,\strut $15$,\strut $20$},
        ymin=0, ymax=20,
        xmin =0, xmax=50,
        grid=major, 
        grid style={dashed,gray}, 
        ylabel near ticks,
        xlabel near ticks,
        xlabel= NVF (\%), 
        ylabel= Accuracy Loss (\%),
        legend columns = 3,
        legend style={at={(1,1)},anchor=north},
        legend style={draw=black, at={(0.5,1)}, text opacity = 1,row sep=0pt, font=\fontsize{6}{4}\selectfont},
         x tick label style={rotate=0,anchor=north},
        ]
        \addplot [smooth, black, thick ,mark=square , mark size=2pt] table [x=NVF, y=Unprotected, col sep=comma] {images/quantization-lenet5-acc-loss.csv};
        \addplot [smooth, blue, thick ,mark=o , mark size=2pt] table [x=NVF, y=Proposed, col sep=comma] {images/quantization-lenet5-acc-loss.csv};
         \addplot [smooth, red, thick,mark=triangle*,mark size=2pt] table [x=NVF, y=TMR, col sep=comma] {images/quantization-lenet5-acc-loss.csv};
    
        \legend{Unprotected, Proposed, TMR}
      \end{axis}
    \end{tikzpicture}

&

\begin{tikzpicture}
\pgfplotsset{
  x ticks with fixed point/.style={
      xticklabel={
        \pgfkeys{/pgf/fpu=true}
        \pgfmathparse{exp(\tick)}%
        \pgfmathprintnumber[fixed relative, precision=3]{\pgfmathresult}
        \pgfkeys{/pgf/fpu=false}
      }
  },
  y ticks with fixed point/.style={
      yticklabel={
        \pgfkeys{/pgf/fpu=true}
        \pgfmathparse{exp(\tick)}%
        \pgfmathprintnumber[fixed relative, precision=3]{\pgfmathresult}
        \pgfkeys{/pgf/fpu=false}
      }
  }
}
  \begin{axis}[title=AlexNet (CIFAR-10),
        width=0.35\textwidth,
        height=0.5\columnwidth,
        scaled x ticks = false,
        scaled y ticks = false,
        xtick={0,5,10,15,20,25,30,35,40,45,50}, 
        ytick={0,5,10,15,20},
        xticklabels = {\strut  $0$,\strut $5$,\strut $10$,\strut $15$,\strut $20$,\strut $25$, \strut$30$,\strut$35$,\strut$40$,\strut$45$, \strut$50$},
        yticklabels = {\strut $0$, \strut $5$,\strut $10$,\strut $15$,\strut $20$},
        ymin=0, ymax=20,
        xmin =0, xmax=50,
        grid=major, 
        grid style={dashed,gray}, 
        ylabel near ticks,
        xlabel near ticks,
        xlabel= NVF (\%), 
        ylabel= Accuracy Loss (\%),
        legend columns = 3,
        legend style={at={(1,1)},anchor=north},
        legend style={draw=black, at={(0.5,1)}, text opacity = 1,row sep=0pt, font=\fontsize{6}{4}\selectfont},
         x tick label style={rotate=0,anchor=north},
        ]
        \addplot [smooth, black, thick ,mark=square , mark size=2pt] table [x=NVF, y=Unprotected, col sep=comma] {images/quantization-alexnet-acc-loss.csv};
        \addplot [smooth, blue, thick ,mark=o , mark size=2pt] table [x=NVF, y=Proposed, col sep=comma] {images/quantization-alexnet-acc-loss.csv};
         \addplot [smooth, red, thick,mark=triangle*,mark size=2pt] table [x=NVF, y=TMR, col sep=comma] {images/quantization-alexnet-acc-loss.csv};
    
        \legend{Unprotected, Proposed, TMR}
      \end{axis}
    \end{tikzpicture}

\\

\;\;\;\;\;\;\;\;\;\;\;\;      \textbf{(a)} &   \;\;\;\;\;\;\;\;\;\;\;\;      \textbf{(b)} &   \;\;\;\;\;\;\;\;\;\;\;\;      \textbf{(c)} 

\\

\begin{tikzpicture}
\pgfplotsset{
  x ticks with fixed point/.style={
      xticklabel={
        \pgfkeys{/pgf/fpu=true}
        \pgfmathparse{exp(\tick)}%
        \pgfmathprintnumber[fixed relative, precision=3]{\pgfmathresult}
        \pgfkeys{/pgf/fpu=false}
      }
  },
  y ticks with fixed point/.style={
      yticklabel={
        \pgfkeys{/pgf/fpu=true}
        \pgfmathparse{exp(\tick)}%
        \pgfmathprintnumber[fixed relative, precision=3]{\pgfmathresult}
        \pgfkeys{/pgf/fpu=false}
      }
  }
}
  \begin{axis}[
        ybar = .01cm,
        bar width = 3.5pt,
        width=0.35\textwidth,
        height=0.5\columnwidth,
        scaled x ticks = false,
        scaled y ticks = false,
        xtick={0,5,10,15,20,25,30,35,40,45,50}, 
        ytick={0,5,10,15,20,25},
        xticklabels = {\strut  $0$,\strut $5$,\strut $10$,\strut $15$,\strut $20$,\strut $25$, \strut$30$,\strut$35$,\strut$40$,\strut$45$, \strut$50$},
        yticklabels = {\strut $0$, \strut $5$,\strut $10$,\strut $15$,\strut $20$, \strut $25$},
        ymin=0, ymax=25,
        xmin =0, xmax=50,
        ylabel near ticks,
        xlabel near ticks,
        xlabel= NVF (\%), 
        ylabel= Critical Faults (\%),
        legend columns = 3,
        legend style={at={(1,1)},anchor=north},
        legend style={draw=black, at={(0.5,1)}, text opacity = 1,row sep=0pt, font=\fontsize{6}{4}\selectfont},
         x tick label style={rotate=0,anchor=north},
        ]

        \addplot [draw=black,fill=yellow] coordinates {(0,20.54) (5,20.54) (10,20.54) (15,20.54) (20,20.54) (25,20.54) (30,20.54) (35,20.54) (40,20.54) (45,20.54) (50,20.54)};
        \addplot [draw=black,fill=blue] coordinates {(0,0.26) (5,0.75) (10,4.1) (15,7.05) (20,11.11) (25,15.32) (30,16.15) (35,18.7) (40,19.44) (45,20.54) (50,20.54)};
        \addplot [draw=black,fill=red] coordinates {(0,0) (5,0.67) (10,4.05) (15,6.86) (20,11.41) (25,14.7) (30,16.37) (35,18.44) (40,19.72) (45,20.25) (50,20.54)};

        \legend{Unprotected, Proposed, TMR}
      \end{axis}
    \end{tikzpicture}

&

\begin{tikzpicture}
\pgfplotsset{
  x ticks with fixed point/.style={
      xticklabel={
        \pgfkeys{/pgf/fpu=true}
        \pgfmathparse{exp(\tick)}%
        \pgfmathprintnumber[fixed relative, precision=3]{\pgfmathresult}
        \pgfkeys{/pgf/fpu=false}
      }
  },
  y ticks with fixed point/.style={
      yticklabel={
        \pgfkeys{/pgf/fpu=true}
        \pgfmathparse{exp(\tick)}%
        \pgfmathprintnumber[fixed relative, precision=3]{\pgfmathresult}
        \pgfkeys{/pgf/fpu=false}
      }
  }
}
  \begin{axis}[
        ybar = .01cm,
        bar width = 3.5pt,
        width=0.35\textwidth,
        height=0.5\columnwidth,
        scaled x ticks = false,
        scaled y ticks = false,
        xtick={0,5,10,15,20,25,30,35,40,45,50}, 
        ytick={0,5,10,15,20,25},
        xticklabels = {\strut  $0$,\strut $5$,\strut $10$,\strut $15$,\strut $20$,\strut $25$, \strut$30$,\strut$35$,\strut$40$,\strut$45$, \strut$50$},
        yticklabels = {\strut $0$, \strut $5$,\strut $10$,\strut $15$,\strut $20$, \strut $25$},
        ymin=0, ymax=25,
        xmin =0, xmax=50,
        ylabel near ticks,
        xlabel near ticks,
        xlabel= NVF (\%), 
        ylabel= Critical Faults (\%),
        legend columns = 3,
        legend style={at={(1,1)},anchor=north},
        legend style={draw=black, at={(0.5,1)}, text opacity = 1,row sep=0pt, font=\fontsize{6}{4}\selectfont},
         x tick label style={rotate=0,anchor=north},
        ]

        \addplot [draw=black,fill=yellow] coordinates {(0,17.84) (5,17.84) (10,17.84) (15,17.84) (20,17.84) (25,17.84) (30,17.84) (35,17.84) (40,17.84) (45,17.84) (50,17.84)};
        \addplot [draw=black,fill=blue] coordinates {(0,0.55) (5,0.61) (10,3.24) (15,5.7) (20,6.44) (25,13.04) (30,17.24) (35,17.84) (40,17.84) (45,17.84) (50,17.84)};
        \addplot [draw=black,fill=red] coordinates {(0,0) (5,0.07) (10,2.63) (15,5.32) (20,6.3) (25,13.08) (30,16.37) (35,17.84) (40,17.84) (45,17.84) (50,17.84)};

        \legend{Unprotected, Proposed, TMR}
      \end{axis}
    \end{tikzpicture}

&

\begin{tikzpicture}
\pgfplotsset{
  x ticks with fixed point/.style={
      xticklabel={
        \pgfkeys{/pgf/fpu=true}
        \pgfmathparse{exp(\tick)}%
        \pgfmathprintnumber[fixed relative, precision=3]{\pgfmathresult}
        \pgfkeys{/pgf/fpu=false}
      }
  },
  y ticks with fixed point/.style={
      yticklabel={
        \pgfkeys{/pgf/fpu=true}
        \pgfmathparse{exp(\tick)}%
        \pgfmathprintnumber[fixed relative, precision=3]{\pgfmathresult}
        \pgfkeys{/pgf/fpu=false}
      }
  }
}
  \begin{axis}[
        ybar = .01cm,
        bar width = 3.5pt,
        width=0.35\textwidth,
        height=0.5\columnwidth,
        scaled x ticks = false,
        scaled y ticks = false,
        xtick={0,5,10,15,20,25,30,35,40,45,50}, 
        ytick={0,5,10,15,20,25},
        xticklabels = {\strut  $0$,\strut $5$,\strut $10$,\strut $15$,\strut $20$,\strut $25$, \strut$30$,\strut$35$,\strut$40$,\strut$45$, \strut$50$},
        yticklabels = {\strut $0$, \strut $5$,\strut $10$,\strut $15$,\strut $20$, \strut $25$},
        ymin=0, ymax=25,
        xmin =0, xmax=50,
        ylabel near ticks,
        xlabel near ticks,
        xlabel= NVF (\%), 
        ylabel= Critical Faults (\%),
        legend columns = 3,
        legend style={at={(1,1)},anchor=north},
        legend style={draw=black, at={(0.5,1)}, text opacity = 1,row sep=0pt, font=\fontsize{6}{4}\selectfont},
         x tick label style={rotate=0,anchor=north},
        ]

        \addplot [draw=black,fill=yellow] coordinates {(0,16.5) (5,16.5) (10,16.5) (15,16.5) (20,16.5) (25,16.5) (30,16.5) (35,16.5) (40,16.5) (45,16.5) (50,16.5)};
        \addplot [draw=black,fill=blue] coordinates {(0,0.11) (5,1.41) (10,4.74) (15,6.84) (20,8.37) (25,10.26) (30,11.82) (35,12.95) (40,14.93) (45,16.46) (50,16.5)};
        \addplot [draw=black,fill=red] coordinates {(0,0) (5,1.24) (10,4.61) (15,6.92) (20,8.48) (25,9.89) (30,11.76) (35,13.31) (40,14.81) (45,16.42) (50,16.5)};

        \legend{Unprotected, Proposed, TMR}
      \end{axis}
    \end{tikzpicture}

\\
\;\;\;\;\;\;\;\;\;\;\;\;      \textbf{(d)} &   \;\;\;\;\;\;\;\;\;\;\;\;      \textbf{(e)} &   \;\;\;\;\;\;\;\;\;\;\;\;      \textbf{(f)}

\\

\begin{tikzpicture}
\pgfplotsset{
  x ticks with fixed point/.style={
      xticklabel={
        \pgfkeys{/pgf/fpu=true}
        \pgfmathparse{exp(\tick)}%
        \pgfmathprintnumber[fixed relative, precision=3]{\pgfmathresult}
        \pgfkeys{/pgf/fpu=false}
      }
  },
  y ticks with fixed point/.style={
      yticklabel={
        \pgfkeys{/pgf/fpu=true}
        \pgfmathparse{exp(\tick)}%
        \pgfmathprintnumber[fixed relative, precision=3]{\pgfmathresult}
        \pgfkeys{/pgf/fpu=false}
      }
  }
}
  \begin{axis}[
        width=0.35\textwidth,
        height=0.5\columnwidth,
        scaled x ticks = false,
        scaled y ticks = false,
        xtick={0,5,10,15,20,25,30,35,40,45,50}, 
        ytick={0,2000, 4000, 6000, 8000, 10000,12000},
        xticklabels = {\strut  $0$,\strut $5$,\strut $10$,\strut $15$,\strut $20$,\strut $25$, \strut$30$,\strut$35$,\strut$40$,\strut$45$, \strut$50$},
        yticklabels = {\strut $0$, \strut $2.0E3$,\strut $4.0E3$,\strut $6.0E3$,\strut $8.0E3$,\strut $1.0E4$,\strut $1.2E4$},
        ymin=0, ymax=12000,
        xmin =0, xmax=50,
        grid=major, 
        grid style={dashed,gray}, 
        ylabel near ticks,
        xlabel near ticks,
        xlabel= NVF (\%), 
        ylabel= \# Neurons,
        legend columns = 3,
        legend style={at={(1,1)},anchor=north},
        legend style={draw=black, at={(0.5,1)}, text opacity = 1,row sep=0pt, font=\fontsize{6}{4}\selectfont},
         x tick label style={rotate=0,anchor=north},
        ]
        \addplot [smooth, black, thick ,mark=square , mark size=2pt] table [x=NVF, y=Unprotected, col sep=comma] {images/quantization-mlp7-neurons-count.csv};
        \addplot [smooth, blue, thick ,mark=o , mark size=2pt] table [x=NVF, y=Proposed, col sep=comma] {images/quantization-mlp7-neurons-count.csv};
        \addplot [smooth, red, thick,mark=triangle*,mark size=2pt] table [x=NVF, y=TMR, col sep=comma] {images/quantization-mlp7-neurons-count.csv};
    
        \legend{Unprotected, Proposed, TMR}
      \end{axis}
    \end{tikzpicture}

&

\begin{tikzpicture}
\pgfplotsset{
  x ticks with fixed point/.style={
      xticklabel={
        \pgfkeys{/pgf/fpu=true}
        \pgfmathparse{exp(\tick)}%
        \pgfmathprintnumber[fixed relative, precision=3]{\pgfmathresult}
        \pgfkeys{/pgf/fpu=false}
      }
  },
  y ticks with fixed point/.style={
      yticklabel={
        \pgfkeys{/pgf/fpu=true}
        \pgfmathparse{exp(\tick)}%
        \pgfmathprintnumber[fixed relative, precision=3]{\pgfmathresult}
        \pgfkeys{/pgf/fpu=false}
      }
  }
}
  \begin{axis}[
        width=0.35\textwidth,
        height=0.5\columnwidth,
        scaled x ticks = false,
        scaled y ticks = false,
        xtick={0,5,10,15,20,25,30,35,40,45,50}, 
        ytick={0,3000, 6000,9000,12000,15000,18000},
        xticklabels = {\strut  $0$,\strut $5$,\strut $10$,\strut $15$,\strut $20$,\strut $25$, \strut$30$,\strut$35$,\strut$40$,\strut$45$, \strut$50$},
        yticklabels = {\strut $0$, \strut $3.0E3$,\strut $6.0E3$,\strut $9.0E3$,\strut $1.2E4$,$1.5E4$,$1.8E4$ },
        ymin=0, ymax=18000,
        xmin =0, xmax=50,
        grid=major, 
        grid style={dashed,gray}, 
        ylabel near ticks,
        xlabel near ticks,
        xlabel= NVF (\%), 
        ylabel= \# Neurons,
        legend columns = 3,
        legend style={at={(1,1)},anchor=north},
        legend style={draw=black, at={(0.5,1)}, text opacity = 1,row sep=0pt, font=\fontsize{6}{4}\selectfont},
         x tick label style={rotate=0,anchor=north},
        ]
        \addplot [smooth, black, thick ,mark=square , mark size=2pt] table [x=NVF, y=Unprotected, col sep=comma] {images/quantization-lenet5-neurons-count.csv};
        \addplot [smooth, blue, thick ,mark=o , mark size=2pt] table [x=NVF, y=Proposed, col sep=comma] {images/quantization-lenet5-neurons-count.csv};
         \addplot [smooth, red, thick,mark=triangle*,mark size=2pt] table [x=NVF, y=TMR, col sep=comma] {images/quantization-lenet5-neurons-count.csv};
    
        \legend{Unprotected, Proposed, TMR}
      \end{axis}
    \end{tikzpicture}

&

\begin{tikzpicture}
\pgfplotsset{
  x ticks with fixed point/.style={
      xticklabel={
        \pgfkeys{/pgf/fpu=true}
        \pgfmathparse{exp(\tick)}%
        \pgfmathprintnumber[fixed relative, precision=3]{\pgfmathresult}
        \pgfkeys{/pgf/fpu=false}
      }
  },
  y ticks with fixed point/.style={
      yticklabel={
        \pgfkeys{/pgf/fpu=true}
        \pgfmathparse{exp(\tick)}%
        \pgfmathprintnumber[fixed relative, precision=3]{\pgfmathresult}
        \pgfkeys{/pgf/fpu=false}
      }
  }
}
  \begin{axis}[
        width=0.35\textwidth,
        height=0.5\columnwidth,
        scaled x ticks = false,
        scaled y ticks = false,
        xtick={0,5,10,15,20,25,30,35,40,45,50}, 
        ytick={50000,100000,150000,200000,250000,300000,350000},
        xticklabels = {\strut  $0$,\strut $5$,\strut $10$,\strut $15$,\strut $20$,\strut $25$, \strut$30$,\strut$35$,\strut$40$,\strut$45$, \strut$50$},
        yticklabels = {\strut $5.0E4$,\strut $1.0E5$,\strut $1.5E5$,\strut $2.0E5$,\strut $2.5E5$,\strut $3.0E5$,\strut $3.5E5$, },
        ymin=50000, ymax=350000,
        xmin =0, xmax=50,
        grid=major, 
        grid style={dashed,gray}, 
        ylabel near ticks,
        xlabel near ticks,
        xlabel= NVF (\%), 
        ylabel= \# Neurons,
        legend columns = 3,
        legend style={at={(1,1)},anchor=north},
        legend style={draw=black, at={(0.5,1)}, text opacity = 1,row sep=0pt, font=\fontsize{6}{4}\selectfont},
         x tick label style={rotate=0,anchor=north},
        ]
        \addplot [smooth, black, thick ,mark=square , mark size=2pt] table [x=NVF, y=Unprotected, col sep=comma] {images/quantization-alexnet-neurons-count.csv};
        \addplot [smooth, blue, thick ,mark=o , mark size=2pt] table [x=NVF, y=Proposed, col sep=comma] {images/quantization-alexnet-neurons-count.csv};
         \addplot [smooth, red, thick,mark=triangle*,mark size=2pt] table [x=NVF, y=TMR, col sep=comma] {images/quantization-alexnet-neurons-count.csv};
    
        \legend{Unprotected, Proposed, TMR}
      \end{axis}
    \end{tikzpicture}

\\

\;\;\;\;\;\;\;\;\;\;\;\;      \textbf{(g)} &   \;\;\;\;\;\;\;\;\;\;\;\;      \textbf{(h)} &   \;\;\;\;\;\;\;\;\;\;\;\;      \textbf{(i)}

\end{tabular}}

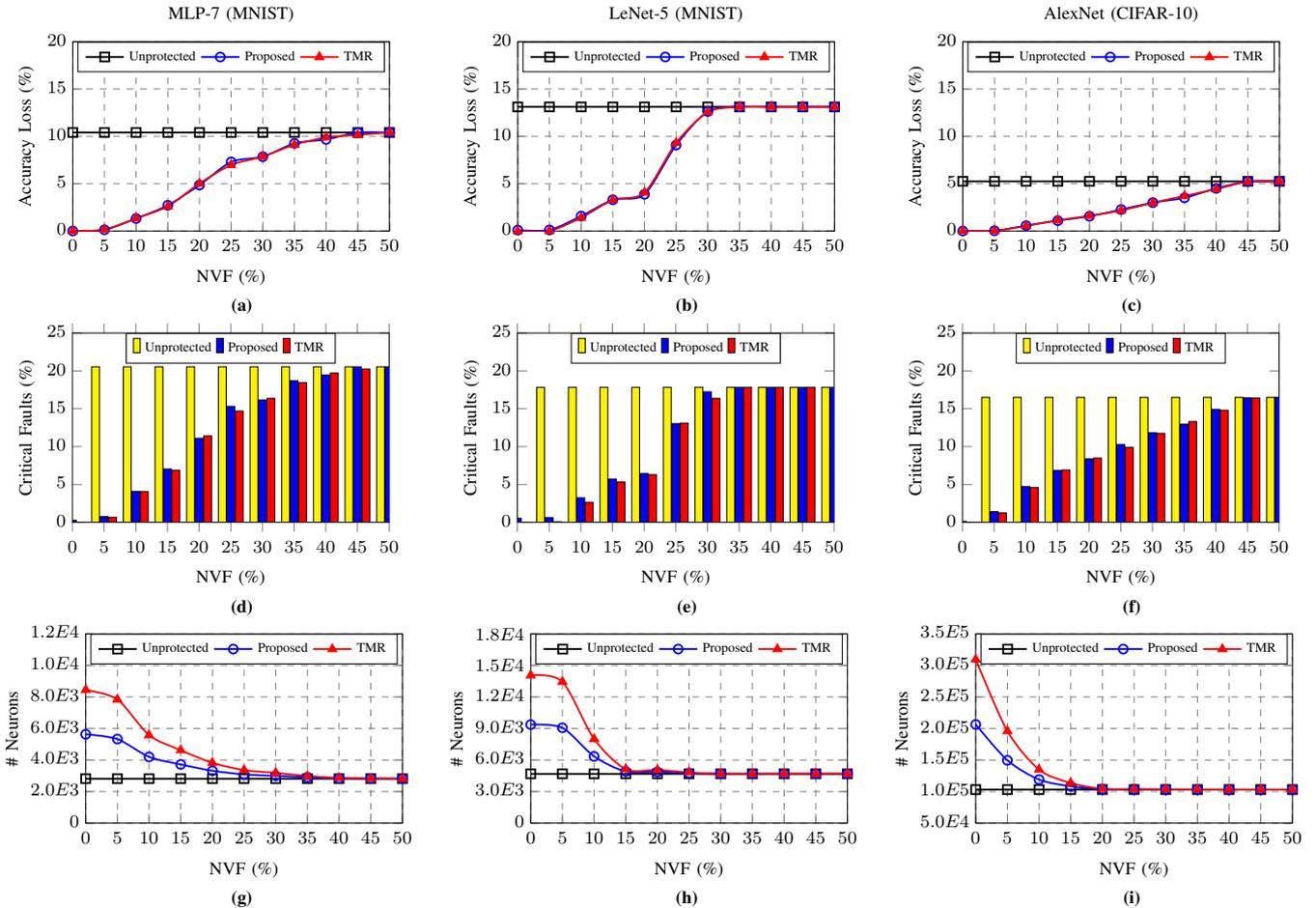
\captionof{figure}{QNNs comparison in terms of accuracy loss (a-c), critical faults (d-f), and network size (g-i) under different levels of protection: unprotected, proposed protection, and TMR, considering different thresholds for NVF from 0\% to 50\%.}
\label{fig:experiments:acc}
\end{table*}

Fig. \ref{fig:experiments:acc} illustrates the experimental results of accuracy loss (a-c) and critical faults (d-f) of the proposed resilience enhancement and TMR over different NVF thresholds for the QNNs. The results show how critical neurons are effectively selected and protected by the proposed method. As shown, all results of protecting QNNs by the proposed method are very close to those of selective TMR-based protection. Furthermore, Fig. \ref{fig:experiments:acc}-(g-i) shows that the QNNs' size (as measured by the number of neurons in each network) using the proposed protection is remarkably smaller than that of the TMR-based protected networks, resulting in half the overhead due to employing one additional neuron for correction instead of two. 

Assuming a constraint on the accuracy loss to be less than 5\% in Fig. \ref{fig:experiments:acc}, a common NVF for all three QNNs can be considered as 20\% in which the accuracy loss is 4.86\%, 3.88\%, and 1.56\% in the QNNs protected by the proposed method that is 2.14x, 3.38x, and 3.36x less than the unprotected QNNs, respectively. Regarding Table \ref{tab:res-nvf-exp}, the resilience analysis suggests protecting 17.86\% of neurons in MLP-7, 3.99\% of neurons in LeNet-5, and 0.6\% of neurons in AlexNet, respectively. The proposed protection mechanism results in 1.85x, 2.78x, and 1.97x fewer critical faults than unprotected QNNs in the MLP-7, LeNet-5, and AlexNet, respectively. 

The proposed neuron splitting and correction method leverages only two neurons (one additional) for correcting faults, whereas TMR requires three neurons (two additional) to perform fault detection and correction. As a result, the overhead of the proposed method is significantly lower than that of TMR, while providing similar resilience. According to Table \ref{tab:res-nvf-exp}, to protect QNNs with an NVF of 20\% using TMR, quantized MLP-7, LeNet-5, and AlexNet require 3,822, 5,058, and 104,412 neurons, respectively, whereas the proposed method requires only 3,319, 4,871, and 103,790 neurons, respectively. Therefore, the proposed method reduces the overall size of QNNs by 15.15\%, 3.84\%, and 0.6\% compared to TMR-based protection, which impacts the memory footprint and execution time of the accelerator accordingly.

\section{Conclusion}  \label{sec:conclusion}

This paper proposes a QNN fault resilience enhancement method. It is achieved by a fault resilience analysis method for QNNs based on the computation of the vulnerability factor for all neurons of a QNN. A neuron splitting method is introduced to modify the network in a way that the critical neurons selected by the resilience analysis are split into two halves. This method enables us to design a Lightweight Correction Unit (LCU) within the accelerator without redesigning its computational parts. The results indicate that the proposed method significantly enhances the fault resiliency of QNNs, matching that of selective TMR methods, but with half the overhead. It means that the proposed method can improve fault resilience in QNNs, making them more reliable for safety-critical applications.

\bibliographystyle{IEEEtran}
\bibliography{refs.bib}

\begin{thebibliography}{10}
\providecommand{\url}[1]{#1}
\csname url@samestyle\endcsname
\providecommand{\newblock}{\relax}
\providecommand{\bibinfo}[2]{#2}
\providecommand{\BIBentrySTDinterwordspacing}{\spaceskip=0pt\relax}
\providecommand{\BIBentryALTinterwordstretchfactor}{4}
\providecommand{\BIBentryALTinterwordspacing}{\spaceskip=\fontdimen2\font plus
\BIBentryALTinterwordstretchfactor\fontdimen3\font minus
  \fontdimen4\font\relax}
\providecommand{\BIBforeignlanguage}[2]{{%
\expandafter\ifx\csname l@#1\endcsname\relax
\typeout{** WARNING: IEEEtran.bst: No hyphenation pattern has been}%
\typeout{** loaded for the language `#1'. Using the pattern for}%
\typeout{** the default language instead.}%
\else
\language=\csname l@#1\endcsname
\fi
#2}}
\providecommand{\BIBdecl}{\relax}
\BIBdecl

\bibitem{silver2017mastering}
D.~Silver \emph{et~al.}, ``Mastering the game of go without human knowledge,''
  \emph{nature}, vol. 550, no. 7676, pp. 354--359, 2017.

\bibitem{mozaffari2020deep}
S.~Mozaffari \emph{et~al.}, ``Deep learning-based vehicle behavior prediction
  for autonomous driving applications: A review,'' \emph{IEEE T-ITS}, 2020.

\bibitem{ibrahim2020soft}
Y.~Ibrahim \emph{et~al.}, ``Soft errors in dnn accelerators: A comprehensive
  review,'' \emph{Microelectronics Reliability}, vol. 115, p. 113969, 2020.

\bibitem{shafique2020robust}
M.~Shafique \emph{et~al.}, ``Robust machine learning systems: Challenges,
  current trends, perspectives, and the road ahead,'' \emph{IEEE Design \&
  Test}, vol.~37, no.~2, pp. 30--57, 2020.

\bibitem{zahid2020fat}
U.~Zahid \emph{et~al.}, ``Fat: Training neural networks for reliable inference
  under hardware faults,'' in \emph{2020 IEEE International Test Conference
  (ITC)}.\hskip 1em plus 0.5em minus 0.4em\relax IEEE, 2020, pp. 1--10.

\bibitem{khoshavi2020fiji}
N.~Khoshavi \emph{et~al.}, ``Fiji-fin: A fault injection framework on quantized
  neural network inference accelerator,'' in \emph{2020 19th IEEE International
  Conference on Machine Learning and Applications (ICMLA)}.\hskip 1em plus
  0.5em minus 0.4em\relax IEEE, 2020, pp. 1139--1144.

\bibitem{mittal2020survey}
S.~Mittal, ``A survey on modeling and improving reliability of dnn algorithms
  and accelerators,'' \emph{Journal of Systems Architecture}, vol. 104, p.
  101689, 2020.

\bibitem{mahmoud2020hardnn}
A.~Mahmoud \emph{et~al.}, ``Hardnn: Feature map vulnerability evaluation in
  cnns,'' \emph{arXiv preprint arXiv:2002.09786}, 2020.

\bibitem{schorn2018accurate}
C.~Schorn and other, ``Accurate neuron resilience prediction for a flexible
  reliability management in neural network accelerators,'' in \emph{2018
  DATE}.\hskip 1em plus 0.5em minus 0.4em\relax IEEE, 2018, pp. 979--984.

\bibitem{schorn2019efficient}
C.~Schorn \emph{et~al.}, ``An efficient bit-flip resilience optimization method
  for deep neural networks,'' in \emph{2019 DATE}.\hskip 1em plus 0.5em minus
  0.4em\relax IEEE, 2019, pp. 1507--1512.

\bibitem{ruospo2021reliability}
A.~Ruospo and E.~Sanchez, ``On the reliability assessment of artificial neural
  networks running on ai-oriented mpsocs,'' \emph{Applied Sciences}, vol.~11,
  no.~14, p. 6455, 2021.

\bibitem{abdullah2020salvagednn}
M.~Abdullah~Hanif and M.~Shafique, ``Salvagednn: salvaging deep neural network
  accelerators with permanent faults through saliency-driven fault-aware
  mapping,'' \emph{Philosophical Transactions of the Royal Society A}, vol.
  378, no. 2164, p. 20190164, 2020.

\bibitem{deepvigor}
M.~H. Ahmadilivani \emph{et~al.}, ``Deepvigor: Vulnerability value ranges and
  factors for dnns reliability assessment,'' in \emph{28th IEEE European Test
  Symposium}.\hskip 1em plus 0.5em minus 0.4em\relax In press, 2023.

\bibitem{ozen2020just}
E.~Ozen and A.~Orailoglu, ``Just say zero: Containing critical bit-error
  propagation in deep neural networks with anomalous feature suppression,'' in
  \emph{39th ICCAD}, 2020, pp. 1--9.

\bibitem{david2021tensorflow}
R.~David \emph{et~al.}, ``Tensorflow lite micro: Embedded machine learning for
  tinyml systems,'' \emph{Machine Learning and Systems}, vol.~3, pp. 800--811,
  2021.

\bibitem{leveugle2009statistical}
R.~Leveugle \emph{et~al.}, ``Statistical fault injection: Quantified error and
  confidence,'' in \emph{2009 DATE}.\hskip 1em plus 0.5em minus 0.4em\relax
  IEEE, 2009, pp. 502--506.

\end{thebibliography}

\end{document}